\documentclass[letterpaper]{article}
\usepackage{preprint}
\usepackage[hyphens]{url}
\usepackage{graphicx}
\usepackage{natbib}
\usepackage{caption}
\usepackage{amsmath,amssymb}
\usepackage{booktabs}
\usepackage{multirow}
\usepackage{tabularx}
\usepackage[table]{xcolor}
\usepackage[hidelinks]{hyperref}

\newcommand{\best}[1]{\textbf{#1}}

\title{Constrained Edit Fields for Training-Free Flow Editing}
\author{Jingxuan Kang, Yinsong Wang, Che Liu, Chen Qin\corresponding}
\affiliations{
    Imperial College London\\
    j.kang26@imperial.ac.uk, c.qin15@imperial.ac.uk
}

\begin{document}

\maketitle

\begin{abstract}
Text-guided image editing aims to perform a desired edit while preserving source content unrelated to it. Pretrained rectified-flow models enable training-free editing of real images through modifications to their sampling trajectories. However, responses at locations unrelated to the desired edit can still accumulate along the editing trajectory and become visible in the final result. To overcome this, we propose \textbf{Constrained Edit Fields (CEF)}, which assigns each spatial location a continuous \emph{edit responsibility} that quantifies its relevance to the desired edit. CEF estimates edit responsibility directly from the source image when the relevant content is present. For edits whose target content is absent from the source, CEF first generates an unconstrained proposal to reveal its realized spatial support and then estimates responsibility from that proposal. At each editing step, CEF decomposes the base edit field into prompt-induced and trajectory-induced components, enabling edit responsibility to preserve instruction-relevant updates while suppressing unintended trajectory-induced changes. Evaluated on all 700 PIE-Bench examples, CEF achieves state-of-the-art Structure Distance, background LPIPS, and background MSE with both Stable Diffusion 3.5 Medium and FLUX, while retaining competitive instruction alignment. On Stable Diffusion 3.5 Medium, it reduces these metrics over the previous best results by 10.2\%, 21.2\%, and 48.0\%, respectively.
\end{abstract}

\section{Introduction}
\begin{figure}[t]
\centering
\includegraphics[width=\columnwidth]{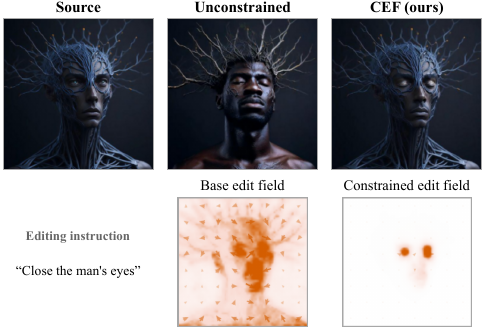}
\caption{Effect of constraining the edit field. The unconstrained edit field introduces updates across the latent grid, closing the subject's eyes while substantially altering his appearance. CEF constrains these updates according to edit responsibility, completing the desired edit while preserving the source appearance.}
\label{fig:teaser}
\end{figure}

Text-guided image editing aims to modify a source image according to an editing instruction while preserving content unrelated to the desired edit. Rectified-flow models~\citep{flowmatching,rectifiedflow} such as SD3 and FLUX have demonstrated strong text-to-image generation capabilities~\citep{sd3,sd35,flux1}. Training-free editing methods leverage these pretrained models to edit real images by modifying their sampling trajectories, without task-specific finetuning~\citep{ftedit,stableflow,rfedit,reflex,flowedit,dnaedit,directedit}. In flow-based editing, the source and target descriptions yield different velocity predictions at each sampling step, and their difference forms the edit field that updates the editing trajectory~\citep{flowedit}. However, this field spans the entire latent grid and can introduce updates at locations unrelated to the desired edit. As shown in Figure~\ref{fig:teaser}, the editor closes the subject's eyes as instructed and substantially alters his appearance and facial structure.

\begin{figure}[t]
\centering
\includegraphics[width=\columnwidth]{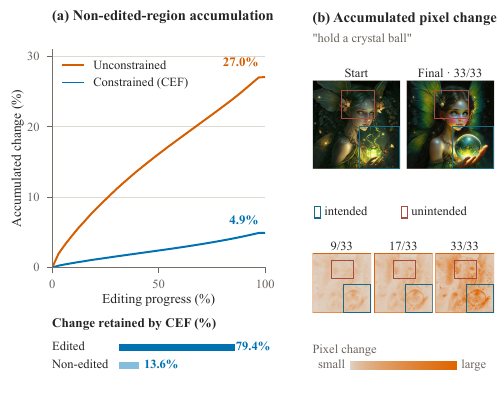}
\caption{Accumulation of unintended changes. (a) Cumulative field displacement outside the edited region, relative to the total unconstrained update inside it. CEF suppresses it while preserving most edited-region updates. (b) Pixel differences along the unconstrained trajectory reveal the intended edit and progressively emerging facial changes.}
\label{fig:accumulation}
\end{figure}

Existing training-free methods improve source preservation by refining editing trajectories~\citep{nulltext,ftedit,rfedit,dnaedit,directedit}, reusing source representations~\citep{prompt2prompt,pnpdiffusion,masactrl,stableflow,reflex,fiaedit}, or providing spatial guidance~\citep{blendeddiffusion,diffedit,kvedit,followyourshape,unieditflow,samflow}. However, the cumulative effect of spatially diffuse edit responses along the trajectory remains insufficiently characterized. Although such responses may be weak at individual steps, repeated integration can amplify them into visible source degradation. As shown in Figure~\ref{fig:accumulation}, the cumulative field displacement outside the edited region reaches 27.0\% of the total unconstrained update within the edited region, and visualizations of the image differences show unrelated facial changes progressively emerging during sampling. These observations demonstrate that weak stepwise responses can accumulate into substantial deviations, highlighting the need to account for each location’s relevance to the desired edit.

We propose \textbf{Constrained Edit Fields (CEF)}, which assigns each spatial location a continuous \emph{edit responsibility} that quantifies how much of its local update is admitted into the constrained edit field. CEF estimates this responsibility from the CLIPSeg responses that localize the content involved in the desired edit~\citep{clipseg}. The editing instruction identifies the relevant content, and visual evidence reveals its realized spatial support. For content observable in the source image, CEF computes responsibility directly from the source response. For content absent from the source, CEF first uses an unconstrained base editor to generate a proposal and estimates responsibility from its response to reveal the realized spatial support. CEF then reuses the sampled noise sequence used to generate the proposal during the constrained edit. The proposal and constrained edit therefore share the sampled noise sequence while following different latent trajectories.

At each editing step, the base edit field compares a target-prompt velocity on the editing trajectory with a source-prompt velocity on the source trajectory, and therefore combines changes in both the prompt and trajectory. CEF exactly decomposes this field into prompt-induced and trajectory-induced changes. The prompt-induced change provides a directional reference for distinguishing the aligned component of the trajectory-induced change. Edit responsibility then modulates their spatial contributions, imposing stronger constraints on the remaining trajectory-induced change where responsibility is low. The resulting constrained field updates the editing trajectory.

Experiments on all 700 PIE-Bench examples~\citep{pnpinversion} using Stable Diffusion 3.5 Medium~\citep{sd3,sd35} and FLUX~\citep{flux1} demonstrate that CEF consistently improves source preservation while maintaining strong editing instruction alignment. CEF establishes new best results for Structure Distance, background LPIPS, and background MSE on both backbones. 

Our contributions are summarized as follows:
\begin{itemize}
\item We show that weak responses at locations unrelated to the desired edit accumulate along the sampling trajectory, and propose edit responsibility to spatially regulate the edit field according to each location’s relevance.
\item We propose CEF, which estimates edit responsibility from visual evidence in the source image or an unconstrained proposal, and exactly decomposes the base edit field into prompt-induced and trajectory-induced changes.
\item Experiments on all 700 PIE-Bench examples using Stable Diffusion 3.5 Medium and FLUX demonstrate that CEF achieves state-of-the-art source preservation while faithfully aligning with the editing instructions.
\end{itemize}

\section{Related Work}
\paragraph{Training-Based Editing.} Training-based approaches learn instruction-conditioned image transformations from editing data. InstructPix2Pix trains a diffusion model on synthetic editing examples to follow natural-language editing instructions~\citep{instructpix2pix}. Emu Edit broadens the range of supported editing operations through multi-task training~\citep{emuedit}, and OmniEdit trains a generalist editor under the supervision of multiple specialist models~\citep{omniedit}. More recently, FLUX.1 Kontext unifies image generation and editing within a flow-matching framework jointly conditioned on a source image and text~\citep{fluxkontext}. Their editing capabilities require editing-specific data and model training.

\begin{figure*}[t]
\centering
\includegraphics[width=\textwidth]{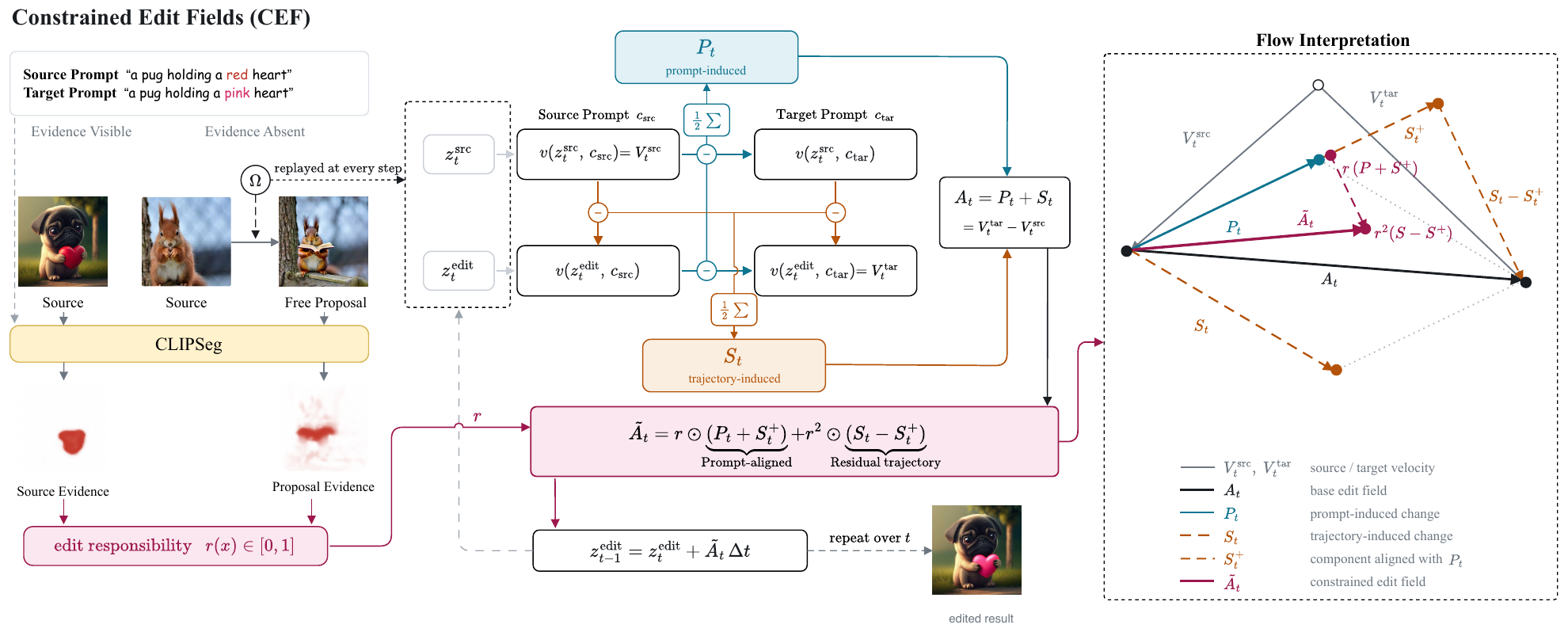}
\caption{Overview of Constrained Edit Fields. Source or proposal evidence localizes the content involved in the edit and yields a continuous edit responsibility.  At each step, the base edit field is exactly decomposed into prompt-induced and trajectory-induced changes.  CEF uses the prompt-induced direction to form prompt-aligned and residual trajectory components, applying responsibility more strongly to the residual to construct the constrained edit field.}
\label{fig:method-overview}
\end{figure*}

\paragraph{Training-Free Editing.} Training-free editing directly leverages pretrained generative models by constructing or modifying their inference trajectories. Early diffusion-based approaches obtain editable representations of the source by altering the sampling initialization or improving the inversion process. SDEdit performs target-conditioned sampling from a partially noised source image~\citep{sdedit}, and Null-text Inversion optimizes unconditional text embeddings to improve real-image inversion and reconstruction~\citep{nulltext}. For rectified-flow models, FTEdit, RF-Inversion, RF-Edit, and FireFlow improve inversion through fixed-point refinement, stochastic inversion, and specialized numerical solvers~\citep{ftedit,rfinversion,rfedit,fireflow}. Another family of methods improves source preservation through representation reuse or trajectory alignment. Stable Flow selectively injects attention features from structure-preserving layers~\citep{stableflow}; ReFlex extracts intermediate source features and adapts attention during editing~\citep{reflex}; DNAEdit aligns the underlying noise~\citep{dnaedit}; and DirectEdit establishes step-level alignment between reconstruction and inversion paths~\citep{directedit}. Recent inversion-free methods construct editing trajectories without explicit inversion. FlowEdit combines source- and target-conditioned velocity fields~\citep{flowedit}; SplitFlow composes prompt-specific sub-flows~\citep{splitflow}; FIA-Edit introduces frequency-interactive attention~\citep{fiaedit}; FlowAlign regularizes the editing trajectory~\citep{flowalign}; and ChordEdit constructs a temporally averaged low-energy control field for one-step editing~\citep{chordedit}. These advances improve reconstruction accuracy, trajectory consistency, and source preservation. But, the spatial extent of the resulting updates remains a separate concern.

\paragraph{Spatially Constrained Editing.} Spatially constrained methods determine which image regions are affected by an edit. Blended Diffusion incorporates a user-provided mask into the denoising process~\citep{blendeddiffusion}, and DiffEdit estimates semantic edit regions by contrasting predictions conditioned on the source and target prompts~\citep{diffedit}. KV-Edit preserves background key--value tokens outside user-specified regions~\citep{kvedit}, and ReFlex uses attention-derived regions for latent blending~\citep{reflex}. Recent approaches further derive spatial cues from the generation process. Follow-Your-Shape constructs a Trajectory Divergence Map from token-wise velocity differences between inversion and editing trajectories to control key--value injection~\citep{followyourshape}. UniEdit-Flow constructs edit regions from per-step velocity differences~\citep{unieditflow}. DirectEdit uses MLLM localization and Segment Anything masks for noise blending, together with attention feature injection~\citep{sam,directedit}. SAM-Flow combines responses from the source and a scout image to construct dynamic editing support and perform source-anchored projection~\citep{samflow}. Across these methods, spatial signals determine where editing updates are applied, with source preservation enforced through blending, feature injection, or trajectory correction. CEF represents visual relevance as continuous edit responsibility and uses it within the base edit field to regulate prompt-induced and trajectory-induced changes.

\section{Method}

Given a source image $x_{\mathrm{src}}$ and source and target prompts $c_{\mathrm{src}}$ and $c_{\mathrm{tar}}$, let $z_t^{\mathrm{src}}$ and $z_t^{\mathrm{edit}}$ denote the source and editing trajectories.  At each editing step, the difference between the velocity predicted on the editing trajectory under the target prompt and that predicted on the source trajectory under the source prompt forms the base edit field.  CEF associates each spatial location with a continuous edit responsibility estimated from visual evidence. CEF then exactly decomposes the base edit field into prompt-induced and trajectory-induced changes at each step, and uses edit responsibility to regulate their contributions to the constrained edit field. Figure~\ref{fig:method-overview} summarizes the complete process.

\begin{figure}[t]
\centering
\includegraphics[width=\columnwidth]{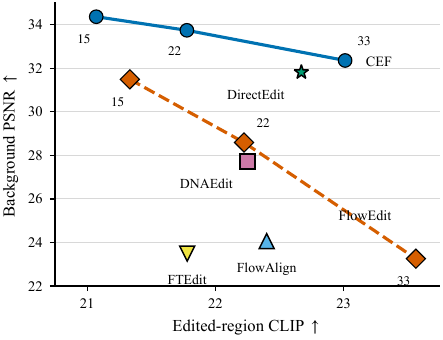}
\caption{CLIP and PSNR on SD3.5. FlowEdit and CEF are evaluated with $n_{\max}\in\{15,22,33\}$.}
\label{fig:frontier}
\end{figure}

\subsection{Edit Responsibility from Visual Evidence}
\label{sec:edit-responsibility}

Preserving source content unrelated to the desired edit requires identifying the image locations involved in the edit. The editing instruction specifies the desired semantic change, and visual evidence localizes the corresponding content in the image. CEF represents each location’s relevance to the desired edit as continuous edit responsibility. This responsibility determines the contribution of the local update to the constrained edit field. Locations with larger responsibility retain more of the edit-induced update. Locations with smaller responsibility are more strongly constrained.

The evidence used to estimate responsibility depends on whether the relevant content is observable in the source image.  Observable content can be localized directly from the source, which provides source evidence.  Content absent from the source provides no corresponding visual evidence.  In this case, CEF first generates an unconstrained proposal $\widetilde x$ in which the target content has been realized and uses it as visual evidence for estimating responsibility.

Given an evidence image $x$ and a query $q$ describing the content involved in the desired edit, CLIPSeg~\citep{clipseg} $g_\phi$ produces a spatial response. CEF converts this response into the edit responsibility
\begin{equation}
r
=
\mathcal R\!\left(g_\phi(x,q)\right),
\label{eq:edit-responsibility}
\end{equation}
where $\mathcal R$ converts the CLIPSeg response into a continuous responsibility field at the spatial resolution of the edit field.  We use $x=x_{\mathrm{src}}$ for source evidence and $x=\widetilde x$ for proposal evidence.  CEF estimates $r$ once from the selected evidence and keeps it fixed throughout the editing trajectory. For proposal evidence, let $\Omega$ denote the sampled noise sequence used to generate $\widetilde x_{\Omega}$, from which CEF estimates $r_{\Omega}$. Since a different sequence may place the target content elsewhere, CEF reuses $\Omega$ during the constrained edit. The proposal and constrained edit therefore share the sampled noise sequence while following different latent trajectories.

\subsection{Decomposing the Edit Field}
\label{sec:field-decomposition}

Let $v_t(z,c)$ denote the guided velocity predicted at step $t$ for latent $z$ under prompt $c$. The base edit field compares predictions that differ in two respects: the target velocity is evaluated on the editing trajectory under the target prompt, whereas the source velocity is evaluated on the source trajectory under the source prompt. Following the differential formulation of training-free flow editing~\citep{flowedit}, the base edit field is
\begin{equation}
A_t
=
v_t\!\left(z_t^{\mathrm{edit}},c_{\mathrm{tar}}\right)
-
v_t\!\left(z_t^{\mathrm{src}},c_{\mathrm{src}}\right).
\label{eq:base-edit-field}
\end{equation}
The field therefore combines the effect of changing the prompt with the effect of moving from the source trajectory to the editing trajectory.  Therefore, treating it as a single quantity does not distinguish the contributions arising from these two sources. To solve this, CEF proposes to separate the two effects without changing the base edit field.

The four relevant velocity predictions pair the source and editing states with the source and target prompts.  Holding the latent state fixed while changing the prompt isolates a prompt-induced velocity change.  For a fixed state $z$, we write

\begin{equation}
\Delta_t^{\mathrm{prompt}}(z)
=
v_t(z,c_{\mathrm{tar}})
-
v_t(z,c_{\mathrm{src}}).
\label{eq:prompt-difference}
\end{equation}
The prompt-induced velocity change depends on the state at which it is evaluated. CEF evaluates it on both trajectories and averages the results to obtain the prompt-induced change,
\begin{equation}
P_t
=
\frac{1}{2}
\left[
\Delta_t^{\mathrm{prompt}}(z_t^{\mathrm{src}})
+
\Delta_t^{\mathrm{prompt}}(z_t^{\mathrm{edit}})
\right].
\label{eq:prompt-change}
\end{equation}

Conversely, holding the prompt fixed while moving from the source state to the editing state isolates a trajectory-induced velocity change,
\begin{equation}
\Delta_t^{\mathrm{traj}}(c)
=
v_t(z_t^{\mathrm{edit}},c)
-
v_t(z_t^{\mathrm{src}},c).
\label{eq:trajectory-difference}
\end{equation}
This change depends on the prompt under which the two states are compared. CEF evaluates it under both prompts and averages the results to obtain the trajectory-induced change,
\begin{equation}
S_t
=
\frac{1}{2}
\left[
\Delta_t^{\mathrm{traj}}(c_{\mathrm{src}})
+
\Delta_t^{\mathrm{traj}}(c_{\mathrm{tar}})
\right].
\label{eq:trajectory-change}
\end{equation}

By construction, the two terms form an exact additive decomposition of the base edit field,
\begin{equation}
A_t=P_t+S_t.
\label{eq:edit-field-decomposition}
\end{equation}
Consequently, the decomposition preserves the update prescribed by the base editor while explicitly seperating the prompt-induced and trajectory-induced changes.

\begin{table*}[t]
\centering
\setlength{\tabcolsep}{2.4pt}
\small
\begin{tabular*}{\textwidth}{@{\extracolsep{\fill}}lc|c|cccc|cc@{}}
\toprule
\multirow{2}{*}{\textbf{Method}} &
\multirow{2}{*}{\textbf{Model}} &
\multicolumn{1}{c}{\textbf{Structure}} &
\multicolumn{4}{c}{\textbf{Background Preservation}} &
\multicolumn{2}{c}{\textbf{CLIP Similarity}} \\
\cmidrule(lr){3-3}\cmidrule(lr){4-7}\cmidrule(lr){8-9}
& & \textbf{Dist.$\downarrow$} &
\textbf{PSNR$\uparrow$} & \textbf{LPIPS$\downarrow$} &
\textbf{MSE$\downarrow$} & \textbf{SSIM$\uparrow$} &
\textbf{Whole$\uparrow$} & \textbf{Edited$\uparrow$} \\
\midrule
\textbf{IP2P} & SD1.5 & 58.13 & 20.95 & 159.20 & 230.87 & 76.39 & 23.61 & \best{21.77} \\
\textbf{P2P} & SD1.5 & \best{15.44} & \best{27.52} & \best{59.69} & \best{34.20} & \best{84.41} & \best{24.75} & 21.01 \\
\midrule
\textbf{RF-Inversion} & FLUX & 41.17 & 20.86 & 187.01 & 120.12 & 71.21 & 25.08 & 22.39 \\
\textbf{RFEdit} & FLUX & 25.15 & 24.33 & 121.59 & 56.98 & 82.84 & 25.57 & 22.54 \\
\textbf{FireFlow} & FLUX & 27.40 & 23.11 & 128.46 & 70.75 & 81.30 & \best{26.13} & \best{22.87} \\
\textbf{FlowEdit} & FLUX & 27.83 & 21.96 & 112.15 & 94.94 & 83.40 & 25.26 & 22.60 \\
\textbf{DNAEdit} & FLUX & 16.81 & 25.20 & 86.68 & 48.35 & 87.21 & 24.81 & 22.12 \\
\textbf{DirectEdit (w/o mask)} & FLUX & 21.93 & 24.70 & 102.92 & 56.76 & 85.76 & 25.89 & 22.74 \\
\textbf{DirectEdit} & FLUX & 17.94 & \best{32.63} & 35.45 & 25.05 & 93.49 & 25.39 & 22.45 \\
\textbf{CEF} & FLUX & \best{11.72} & 31.22 & \best{29.22} & \best{16.21} & \best{93.60} & 24.48 & 21.84 \\
\midrule
\textbf{FTEdit} & SD3.5 & 21.06 & 23.49 & 90.25 & 61.78 & 86.23 & 25.21 & 21.78 \\
\textbf{FlowEdit} & SD3.5 & 23.13 & 23.29 & 92.81 & 69.09 & 85.22 & \best{26.71} & \best{23.59} \\
\textbf{FlowAlign} & SD3.5 & 33.49 & 24.06 & 68.91 & 54.29 & 86.33 & 25.64 & 22.40 \\
\textbf{DNAEdit} & SD3.5 & 11.03 & 27.71 & 60.51 & 26.28 & 90.13 & 25.20 & 22.25 \\
\textbf{DirectEdit (w/o mask)} & SD3.5 & 15.23 & 26.18 & 67.28 & 34.00 & 88.75 & 26.13 & 22.50 \\
\textbf{DirectEdit} & SD3.5 & 14.65 & 31.82 & 31.36 & 21.64 & 92.28 & 25.64 & 22.67 \\
\textbf{SAM-Flow} & SD3.5 & 12.20 & 27.82 & 46.24 & 30.25 & 90.66 & 26.15 & 23.12 \\
\textbf{CEF} & SD3.5 & \best{9.91} & \best{32.72} & \best{24.72} & \best{11.26} & \best{92.98} & 25.52 & 22.91 \\
\bottomrule
\end{tabular*}
\caption{Quantitative comparisons on PIE-Bench. Baseline aggregates except SAM-Flow are reproduced from Table~1 of DirectEdit~\citep{directedit}; the SAM-Flow and CEF rows are evaluated by us. Structure distance and LPIPS are reported at $10^3$ scale, MSE at $10^4$ scale, and SSIM at $10^2$ scale. Bold indicates the best result within each backbone.}
\label{tab:main-detailed}
\end{table*}

\begin{figure*}[t]
\centering
\includegraphics[width=0.97\textwidth]{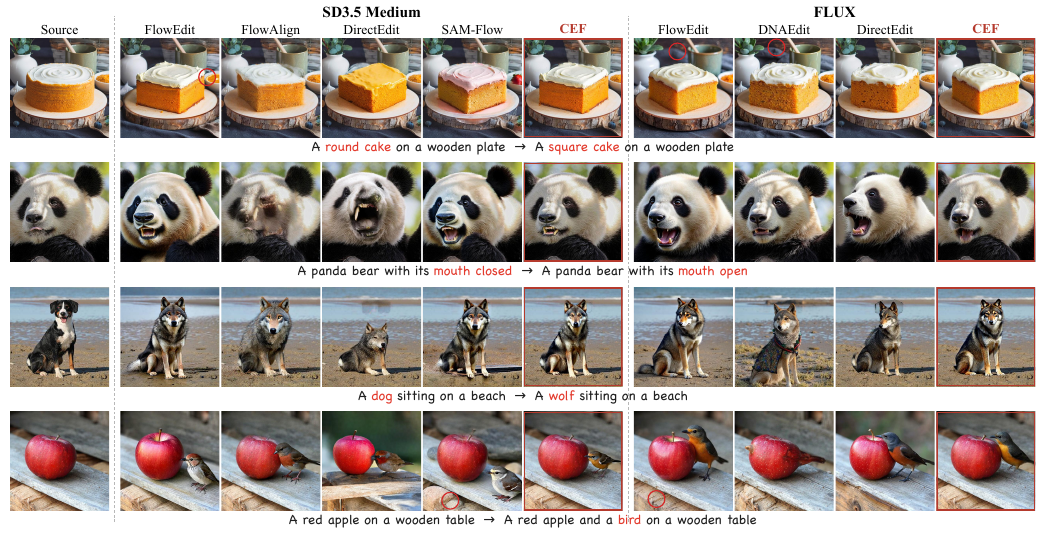}
\caption{Qualitative comparison on PIE-Bench. The same edits are compared under fixed configurations on SD3.5 Medium and FLUX. CEF preserves source appearance while realizing shape changes, attribute edits, object replacements, and object additions across both backbones. Red circles mark incomplete target edits or unintended changes in regions outside the edit.}
\label{fig:qual}
\end{figure*}

\begin{figure}[t]
\centering
\setlength{\tabcolsep}{0pt}
\small
\begin{tabular*}{\columnwidth}{@{\extracolsep{\fill}}l|r|rrrr|rr@{}}
\toprule
\multirow{2}{*}{\textbf{Config.}} &
\multicolumn{1}{c|}{\textbf{Struct.}} &
\multicolumn{4}{c|}{\textbf{Background}} &
\multicolumn{2}{c}{\textbf{CLIP}} \\
\cmidrule(lr){2-2}\cmidrule(lr){3-6}\cmidrule(lr){7-8}
& \textbf{Dist.$\downarrow$} & \textbf{PSNR$\uparrow$} &
\textbf{LPIPS$\downarrow$} & \textbf{MSE$\downarrow$} &
\textbf{SSIM$\uparrow$} & \textbf{Whole$\uparrow$} &
\textbf{Edit$\uparrow$} \\
\midrule
\textbf{w/o $r$} & 10.98 & 32.08 & 26.63 & 13.51 &
92.75 & 25.66 & 23.10 \\
\textbf{w/o prop.} & 12.49 & 31.66 & 32.22 & 18.11 &
92.09 & \best{25.76} & \best{23.14} \\
\textbf{CEF (Ours)} & \best{9.91} & \best{32.72} & \best{24.72} &
\best{11.26} & \best{92.98} & 25.52 & 22.91 \\
\bottomrule
\end{tabular*}
\begingroup
\captionsetup{type=table}
\caption{Component ablation on PIE-Bench with SD3.5 Medium. Each ablation independently removes one component from CEF.}
\label{tab:ablation}
\endgroup
\end{figure}

\subsection{The Constrained Edit Field}
\label{sec:constrained-edit-field}

Decomposing the base field allows CEF to distinguish the part of the trajectory-induced change that follows the prompt-induced direction. At each spatial location, CEF uses $P_t$ as a local directional reference and extracts the positively aligned component of $S_t$,
\begin{equation}
S_t^{+}
=
\left[
\frac{\langle S_t,P_t\rangle}
{\lVert P_t\rVert_2^{2}+\epsilon}
\right]_{+}P_t,
\label{eq:prompt-aligned-change}
\end{equation}
where the inner product and norm act over the channel dimension at each spatial location, $[a]_{+}=\max(a,0)$, and $\epsilon$ provides numerical stability. The positive projection $S_t^{+}$ follows the local direction of $P_t$.  We refer to $P_t+S_t^{+}$ as the \emph{prompt-aligned component} and to $S_t-S_t^{+}$ as the \emph{residual trajectory component}.

CEF uses edit responsibility to regulate these two components differently. It applies $r$ to the prompt-aligned component $P_t+S_t^{+}$ and $r^2$ to the residual trajectory component.  The constrained edit field can be formulated as
\begin{equation}
\widetilde A_t
=
r\odot\left(P_t+S_t^{+}\right)
+
r^{2}\odot\left(S_t-S_t^{+}\right),
\label{eq:constrained-edit-field}
\end{equation}

where $\odot$ denotes the Hadamard product, with $r$ and $r^2$ broadcast across the channel dimension. At any location with $r>0$, the ratio between the coefficients applied to the residual trajectory component and the prompt-aligned component is $r$. As edit responsibility decreases, the residual trajectory component is therefore attenuated more rapidly, changing both the magnitude and composition of the local update rather than uniformly scaling the base edit field.

The construction preserves the two endpoints of edit responsibility.  At a location with $r=1$, the two terms recombine as $P_t+S_t=A_t$; at a location with $r=0$, the constrained update vanishes.

CEF then directly uses the constrained edit field to update the editing trajectory,
\begin{equation}
z_{t-1}^{\mathrm{edit}}
=
z_t^{\mathrm{edit}}
+
\eta_t\widetilde A_t,
\label{eq:constrained-update}
\end{equation}
where $\eta_t$ is the signed interval between consecutive sampling times in the base editor.

\section{Experiments}

\subsection{Experimental Setup}

\paragraph{Evaluation Datasets and Metrics.} We evaluate CEF on all 700 examples in PIE-Bench~\citep{pnpinversion}. We measure alignment with the target prompt using CLIP similarity~\citep{clip} between the prompt and either the complete edited image (CLIP-W) or the annotated edit region (CLIP-E). Source preservation is evaluated using DINO-based Structure Distance~\citep{pnpinversion,dino} over the complete image, together with PSNR, MSE, LPIPS~\citep{lpips}, and SSIM~\citep{ssim} over the region outside the edit. Benchmark masks define the edited region and the remaining image only for evaluation and are not accessed by CEF.

\paragraph{Implementation Details.} We implement CEF using Stable Diffusion 3.5 Medium~\citep{sd3,sd35} and FLUX~\citep{flux1} as the backbone models, while keeping all generative models frozen. For direct comparison, we use the sampling configurations of FlowEdit~\citep{flowedit}: $(T,n_{\max})=(50,33)$ for SD3.5 and $(T,n_{\max})=(28,24)$ for FLUX, with $n_{\mathrm{avg}}=1$ for both. The classifier-free guidance scales~\citep{cfg} for the source and target are $3.5$ and $13.5$ for SD3.5, and $1.5$ and $5.5$ for FLUX. The source prompt, target prompt, and editing instruction determine the evidence used by CEF and the content queried by CLIPSeg. For multiple grounding phrases, CEF takes a pixel-wise maximum over their CLIPSeg responses, bilinearly resizes the combined response to image resolution, and applies per-image min--max normalization to obtain $s\in[0,1]$, in this order. CEF then maps $s$ to image-space responsibility using $\operatorname{clip}(\alpha s^\gamma,0,1)$, where $\alpha$ is the responsibility gain and $\gamma$ is the response exponent. The resulting responsibility map is resized to the spatial resolution of the edit field. SD3.5 uses $(\alpha,\gamma)=(2.3,1.5)$ for source evidence and $(1.0,1.0)$ for proposal evidence; FLUX uses $(4.0,1.0)$ for both. For global edits, we set $r\equiv1$ over the spatial grid, so $\widetilde A_t=A_t$ and CEF reduces to the base editor. Benchmark categories and masks are not used. Experiments are conducted in a Linux container on NVIDIA A800-SXM4 GPUs with 80~GB of memory, using Python 3.11 and PyTorch 2.11 with CUDA 12.8. Responsibility construction and processing details are provided in Appendix~\ref{app:responsibility}. Runtime and memory analyses are provided in Appendix~\ref{app:runtime}.

\paragraph{Comparison Methods.} For quantitative comparison, we use the published PIE-Bench aggregates reported by DirectEdit~\citep{directedit}. They cover diffusion-based editors InstructPix2Pix~\citep{instructpix2pix} and Prompt-to-Prompt~\citep{prompt2prompt}, together with the rectified-flow methods FTEdit~\citep{ftedit}, RF-Inversion~\citep{rfinversion}, RFEdit~\citep{rfedit}, FireFlow~\citep{fireflow}, FlowEdit~\citep{flowedit}, FlowAlign~\citep{flowalign}, and DNAEdit~\citep{dnaedit}. DirectEdit and its mask-free variant are taken from the same report. We run SAM-Flow~\citep{samflow} on all 700 examples using the authors' implementation and evaluate it with the same protocol.

\subsection{Main Results}

Table~\ref{tab:main-detailed} reports results on all 700 PIE-Bench examples. CEF achieves the best Structure Distance and all four metrics measured outside the edited region on SD3.5. Compared with DNAEdit, the previous best Structure Distance result, CEF reduces it from 11.03 to 9.91. Compared with DirectEdit, the previous strongest baseline on the background metrics, CEF reduces LPIPS from 31.36 to 24.72 and MSE from 21.64 to 11.26, while attaining a higher CLIP-E. FlowEdit obtains the highest CLIP scores but performs substantially worse on the preservation metrics. Figure~\ref{fig:frontier} compares CLIP-E and background PSNR for three values of $n_{\max}$. Across these settings, CEF achieves higher PSNR than FlowEdit at similar CLIP-E. The main CEF configuration also has higher PSNR than all SD3.5 baselines in the table. Complete frontier metrics are provided in Appendix~\ref{app:frontier}. On FLUX, CEF also achieves the best Structure Distance, background LPIPS, MSE, and SSIM while retaining competitive CLIP-E, confirming consistent preservation gains across both backbones.

\paragraph{Qualitative comparison.} Figure~\ref{fig:qual} compares four local edits across the two backbones. FlowEdit often completes the requested edit but may also change nearby content or the appearance of the source object, as shown by the cake and bird examples. FlowAlign and DNAEdit sometimes leave the requested edit incomplete or change the pose and scale of the edited object. DirectEdit and SAM-Flow localize the edit more accurately but can still alter the source appearance or introduce local artifacts. Across both backbones, CEF changes the cake geometry while retaining its frosting and surroundings, opens the panda's mouth without altering its facial appearance, transforms the dog while preserving its pose and background, and adds the bird without changing the apple or table. Additional qualitative results are provided in Appendix~\ref{app:qualitative}; failure cases and limitations are discussed in Appendix~\ref{app:failures}.

\subsection{Ablation Studies}

\paragraph{Ablation.} Table~\ref{tab:ablation} evaluates the two components independently. Replacing continuous responsibility with a binary decision increases structure change and background error. Removing proposal evidence causes a larger loss of preservation when the target content is absent from the source image. CLIP-E changes only slightly in both cases. Continuous responsibility and proposal evidence therefore improve preservation while keeping CLIP-E similar.

\begin{figure}[t]
\centering
\includegraphics[width=\columnwidth]{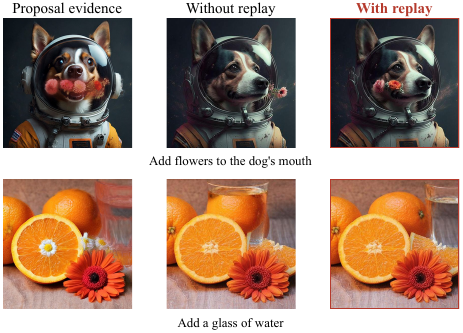}
\caption{Effect of replay. Both final constrained edits are run with the same realization $\Omega$. With replay, responsibility $r_{\Omega}$ is estimated from the proposal generated with $\Omega$, as shown by the red mask in the first column.}
\label{fig:replay-qual}
\end{figure}
\FloatBarrier

\begin{figure}[t]
\centering
\includegraphics[width=\columnwidth]{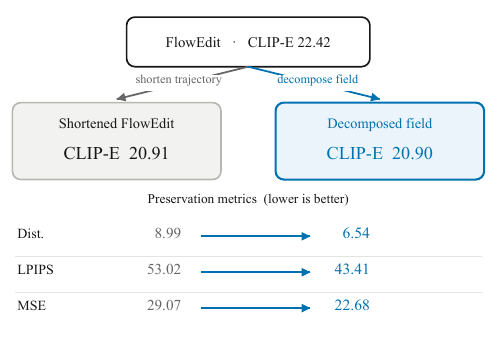}
\caption{Effect of decomposition. The decomposed field gives lower Structure Distance, LPIPS, and MSE.}
\label{fig:decomposition-control}

\setlength{\tabcolsep}{3.2pt}
\small
\begin{tabularx}{\columnwidth}{@{}*{5}{>{\centering\arraybackslash}X}@{}}
\toprule
$\alpha$ & $\gamma$ & \textbf{Dist.$\downarrow$} & \textbf{LPIPS$\downarrow$} & \textbf{CLIP-E$\uparrow$} \\
\midrule
3.5 & 1.0 & 7.69 & 24.35 & \best{20.37} \\
\rowcolor{blue!10} 4.0$^\dagger$ & 1.0$^\dagger$ & 7.98 & 25.06 & 20.34 \\
4.5 & 1.0 & 8.19 & 25.85 & 20.32 \\
4.0 & 0.8 & 8.76 & 27.81 & 20.32 \\
4.0 & 1.2 & \best{7.31} & \best{23.57} & 20.23 \\
\bottomrule
\end{tabularx}
\begingroup
\captionsetup{type=table}
\caption{Sensitivity to $\alpha$ and $\gamma$ on FLUX. Each row changes one parameter from the setting used in the main experiments, using the same 50-example subset. Blue shading marks the main setting, and bold indicates the best result in each column. The main setting was not selected using this subset and is not best in any column.}
\label{tab:hparam-sensitivity}
\endgroup
\end{figure}
\paragraph{Decomposition improves preservation.} To isolate the effect of decomposition, we apply a single fixed coefficient to $S_t$ across all locations and active editing steps. Since $A_t=P_t+S_t$, a coefficient of $1$ on $S_t$ exactly recovers the original FlowEdit field. We set the coefficient to $0.5$ as a simple non-adaptive control; this coefficient is not selected as an optimum. The resulting field $P_t+0.5S_t$ retains all 24 active steps. We then shorten FlowEdit to 20 active steps to match its editing strength. The two settings produce nearly identical CLIP-E values of 20.90 and 20.91. As shown in Figure~\ref{fig:decomposition-control}, the decomposed field reduces Structure Distance from 8.99 to 6.54, LPIPS from 53.02 to 43.41, and MSE from 29.07 to 22.68. Separately attenuating $S_t$ therefore provides better source preservation than reducing the number of active FlowEdit steps with nearly identical CLIP-E.

\paragraph{Effect of replay.} We evaluate replay on all 80 PIE-Bench edits that add an object. For the controlled comparison, both final constrained edits use the same fixed realization $\Omega$. With replay, responsibility $r_{\Omega}$ is estimated from the proposal generated with $\Omega$; without replay, responsibility $r_{\Omega'}$ is estimated from an independently generated proposal with $\Omega'$. This isolates the effect of realization-matched proposal evidence. Figure~\ref{fig:replay-qual} shows that replay makes the final edit more closely retain the location and extent of the added content in the proposal. Across all 80 edits, the overlap between the added-object regions in the proposal and final result increases from 0.708 to 0.794. The paired 95\% bootstrap confidence interval indicates an increase of 4.6 to 12.8 percentage points.

\paragraph{Hyperparameter sensitivity.} Table~\ref{tab:hparam-sensitivity} varies $\alpha$ and $\gamma$ one at a time around the values used in the main FLUX experiments. Across the evaluated values, CLIP-E changes by only 0.14, while Structure Distance and LPIPS change gradually. Larger values of $\gamma$ reduce both preservation errors while CLIP-E remains similar. The main setting, $(\alpha,\gamma)=(4.0,1.0)$, was not selected using this subset and is not best in any reported metric. Complete sensitivity results are provided in Appendix~\ref{app:sensitivity}.

\section{Conclusion}

We propose CEF, a training-free method, to address unintended source changes caused by edit responses accumulating outside the requested region along editing trajectories. CEF estimates continuous edit responsibility from visual evidence, decomposes the base edit field into prompt-induced and trajectory-induced components, and spatially constrains their contributions at each editing step. This design suppresses unrelated changes without uniformly weakening the editing trajectory. Across all 700 PIE-Bench examples, CEF achieves state-of-the-art source preservation with SD3.5 Medium and FLUX while maintaining edit alignment.

\bibliography{main}

\clearpage
\onecolumn
\appendix
\setcounter{secnumdepth}{2}
\renewcommand{\thesection}{\Alph{section}}
\setcounter{table}{0}
\setcounter{figure}{0}
\renewcommand{\thetable}{A\arabic{table}}
\renewcommand{\thefigure}{A\arabic{figure}}
\section{Responsibility Construction and Processing Details}
\label{app:responsibility}

\paragraph{Visual evidence selection.}
CEF derives grounding phrases and selects the visual evidence source from the source description, target description, and editing instruction. When the edit-related content is observable in the source image, CEF estimates edit responsibility from the source image. When the requested target content is generated by the edit, CEF first runs the base editor to produce a proposal and estimates responsibility from the realized target content. For global edits, CEF sets $r\equiv1$. The evidence source and grounding phrases are determined before constrained generation begins.

\paragraph{Responsibility construction.}
For each grounding phrase, the frozen CLIPSeg model~\citep{clipseg} produces an image-space semantic response. CEF combines responses from multiple phrases by a pixel-wise maximum, bilinearly resizes the combined response to image resolution, and applies per-image min--max normalization to obtain $s\in[0,1]$. It maps the semantic response to edit responsibility as
\begin{equation}
r=\operatorname{clip}(\alpha s^\gamma,0,1).
\end{equation}
CEF then resizes the responsibility to the spatial resolution of the latent edit field using area interpolation. The responsibility is computed once and held fixed over all active editing steps. SD3.5~\citep{sd3,sd35} uses $(\alpha,\gamma)=(2.3,1.5)$ for source evidence and $(1.0,1.0)$ for proposal evidence. FLUX~\citep{flux1} uses $(4.0,1.0)$ for both evidence sources.

\paragraph{Proposal evidence and stochastic replay.}
For edits using proposal evidence, CEF first runs the base editor to produce an unconstrained proposal and evaluates the target grounding phrase on the realized proposal. During constrained editing, CEF replays the noise sequence sampled for proposal generation. The proposal and constrained edit share the same noise realization and evolve along their respective latent trajectories. The resulting responsibility corresponds to the target content realized under that noise sequence.

\paragraph{Sampling configurations.}
SD3.5 uses 50 sampling steps, $n_{\max}=33$, $n_{\mathrm{avg}}=1$, and source and target guidance scales of 3.5 and 13.5. FLUX uses 28 sampling steps, $n_{\max}=24$, $n_{\mathrm{avg}}=1$, and source and target guidance scales of 1.5 and 5.5. The parameters of the generative models and visual encoder remain frozen throughout the experiments.

\section{Runtime and Memory Analysis}
\label{app:runtime}

We measure the runtime and peak allocated memory of FlowEdit~\citep{flowedit} and CEF with SD3.5 Medium on a single NVIDIA A800-SXM4-80GB GPU at $512\times512$ resolution with one sample per run. The model and VAE use FP16, with $T=50$, $n_{\max}=33$, and $n_{\mathrm{avg}}=1$. Timing begins with source-image VAE encoding and covers text encoding, the complete editing trajectory, and final VAE decoding. CEF receives a precomputed edit responsibility at the beginning of each timed run.

Each method is measured in two runs. Each run contains five warm-up samples followed by 55 timed samples, giving 110 measurements per method. Table~\ref{tab:app-runtime} reports the mean and standard deviation of runtime and the peak allocated memory recorded by PyTorch.

\begin{table}[ht]
\centering
\small
\setlength{\tabcolsep}{5pt}
\begin{tabular}{lcc}
\toprule
Method & Runtime (s) $\downarrow$ & Peak memory (GiB) $\downarrow$ \\
\midrule
FlowEdit & $3.5736 \pm 0.0023$ & 26.13 \\
CEF & $5.1050 \pm 0.0040$ & 26.23 \\
\bottomrule
\end{tabular}
\caption{Editor runtime and peak allocated memory on SD3.5. Each timed run covers source and prompt encoding, the editing trajectory, and decoding, with a precomputed edit responsibility provided to CEF.}
\label{tab:app-runtime}
\end{table}

Under this configuration, CEF takes $1.4285\times$ the FlowEdit runtime and adds approximately 0.11~GiB of peak allocated memory.

\section{Complete Hyperparameter Sensitivity}
\label{app:sensitivity}

\begin{table}[t]
\centering
\small
\setlength{\tabcolsep}{2.0pt}
\begin{tabular*}{\textwidth}{@{\extracolsep{\fill}}lccrrrrrrr@{}}
\toprule
\textbf{Backbone} & $\boldsymbol{\alpha}_{\mathrm{s}}/\boldsymbol{\alpha}_{\mathrm{p}}$ & $\boldsymbol{\gamma}_{\mathrm{s}}/\boldsymbol{\gamma}_{\mathrm{p}}$ &
\textbf{Dist.$\downarrow$} & \textbf{PSNR$\uparrow$} &
\textbf{LPIPS$\downarrow$} & \textbf{MSE$\downarrow$} &
\textbf{SSIM$\uparrow$} & \textbf{CLIP-W$\uparrow$} &
\textbf{CLIP-E$\uparrow$} \\
\midrule
SD3.5 ($n=50$) & 1.8 / 0.8 & 1.5 / 1.0 & 6.951 & 34.670 & 22.181 & 7.657 & 93.631 & 25.211 & 20.833 \\
SD3.5 ($n=50$) & 2.3$^\dagger$ / 1.0$^\dagger$ & 1.5$^\dagger$ / 1.0$^\dagger$ & 7.538 & 34.155 & 23.223 & 8.640 & 93.525 & 25.511 & 21.067 \\
SD3.5 ($n=50$) & 2.8 / 1.2 & 1.5 / 1.0 & 8.214 & 33.787 & 24.014 & 9.396 & 93.452 & 25.612 & 21.085 \\
SD3.5 ($n=50$) & 2.3 / 1.0 & 1.2 / 0.8 & 7.940 & 33.653 & 24.378 & 9.405 & 93.422 & 25.589 & 21.144 \\
SD3.5 ($n=50$) & 2.3 / 1.0 & 1.8 / 1.2 & 7.214 & 34.498 & 22.591 & 8.138 & 93.586 & 25.270 & 20.919 \\
\midrule
FLUX ($n=50$) & 3.5 & 1.0 & 7.693 & 33.399 & 24.347 & 11.239 & 94.746 & 24.347 & 20.369 \\
FLUX ($n=50$) & 4.0$^\dagger$ & 1.0$^\dagger$ & 7.985 & 33.096 & 25.057 & 11.890 & 94.652 & 24.367 & 20.344 \\
FLUX ($n=50$) & 4.5 & 1.0 & 8.187 & 32.836 & 25.848 & 12.439 & 94.569 & 24.278 & 20.316 \\
FLUX ($n=50$) & 4.0 & 0.8 & 8.761 & 32.041 & 27.812 & 14.029 & 94.306 & 24.343 & 20.319 \\
FLUX ($n=50$) & 4.0 & 1.2 & 7.309 & 33.806 & 23.570 & 10.450 & 94.856 & 24.219 & 20.228 \\
\bottomrule
\end{tabular*}
\caption{Sensitivity to $\alpha$ and $\gamma$ on the same 50-example subset used in Table~\ref{tab:hparam-sensitivity}. Every row aggregates all 50 examples. Daggers mark the configurations used in the main experiments. Structure and LPIPS are reported at $10^3$ scale, MSE at $10^4$ scale, and SSIM at $10^2$ scale.}
\label{tab:app-hparam-sensitivity}
\end{table}

After fixing the main experimental configurations, we evaluate sensitivity to $\alpha$ and $\gamma$ on the fixed 50-example PIE-Bench~\citep{pnpinversion} subset used in Table~\ref{tab:hparam-sensitivity}. Each row reports the aggregate over the complete subset, and the two backbones use the same sample identifiers. We vary one mapping parameter at a time while keeping all other parameters fixed.

These results show that CEF is robust to moderate variations in $\alpha$ and $\gamma$ across both backbones. CLIP-E varies by at most 0.25 on SD3.5 and 0.14 on FLUX, while the preservation metrics remain stable across the evaluated settings.

\section{Complete Results across Preservation and Editing Settings}
\label{app:frontier}

Table~\ref{tab:app-frontier} provides the metrics underlying the SD3.5 tradeoff plot in Figure~\ref{fig:frontier}.

\begin{table}[ht]
\centering
\footnotesize
\setlength{\tabcolsep}{2.0pt}
\begin{tabular*}{\textwidth}{@{\extracolsep{\fill}}lcrrrrrrrr@{}}
\toprule
\textbf{Method} & $\boldsymbol{n}_{\max}$ &
\textbf{Dist.$\downarrow$} & \textbf{PSNR$\uparrow$} &
\textbf{LPIPS$\downarrow$} & \textbf{MSE$\downarrow$} &
\textbf{SSIM$\uparrow$} & \textbf{CLIP-W$\uparrow$} &
\textbf{CLIP-E$\uparrow$} \\
\midrule
FlowEdit & 15 & 3.71 & 31.49 & 31.53 & 12.41 & 92.13 & 24.23 & 21.33 \\
FlowEdit & 22 & 7.60 & 28.59 & 46.14 & 22.59 & 90.37 & 25.33 & 22.22 \\
FlowEdit & 33 & 23.48 & 23.25 & 93.41 & 69.19 & 85.14 & 26.78 & 23.56 \\
\midrule
CEF & 15 & 2.00 & 34.39 & 20.28 & 7.09 & 93.49 & 23.70 & 21.06 \\
CEF & 22 & 3.86 & 33.81 & 21.58 & 8.03 & 93.33 & 24.52 & 21.75 \\
CEF & 33 & 9.91 & 32.72 & 24.72 & 11.26 & 92.98 & 25.52 & 22.91 \\
\bottomrule
\end{tabular*}
\caption{Results for FlowEdit and CEF on all 700 PIE-Bench examples at $n_{\max}\in\{15,22,33\}$. Structure and LPIPS are reported at $10^3$ scale, MSE at $10^4$ scale, and SSIM at $10^2$ scale.}
\label{tab:app-frontier}
\end{table}

\FloatBarrier

\section{Additional Qualitative Results}
\label{app:qualitative}

Figure~\ref{fig:app-qual} provides additional fixed-configuration comparisons beyond the four cases in Figure~\ref{fig:qual}.

\newpage
\begin{center}
\centering
\includegraphics[width=0.99\textwidth,height=0.90\textheight,keepaspectratio]{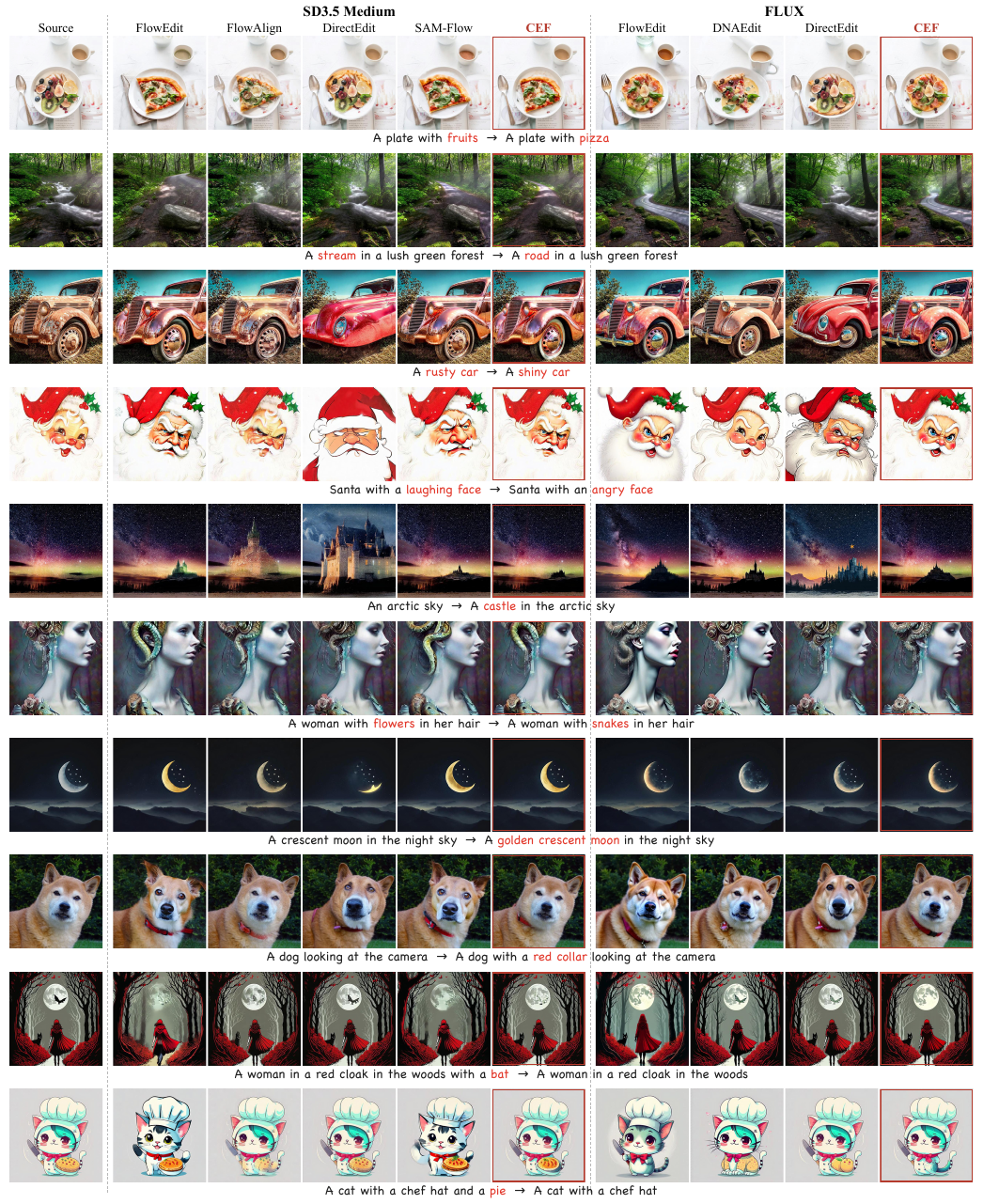}
\captionof{figure}{Additional qualitative comparisons on PIE-Bench. The same edits are compared under the fixed SD3.5 Medium and FLUX configurations used in the main experiments.}
\label{fig:app-qual}
\end{center}
\FloatBarrier

\newpage
\section{Failure Cases and Limitations}
\label{app:failures}

\begin{figure}[ht]
\centering
\includegraphics[width=\textwidth]{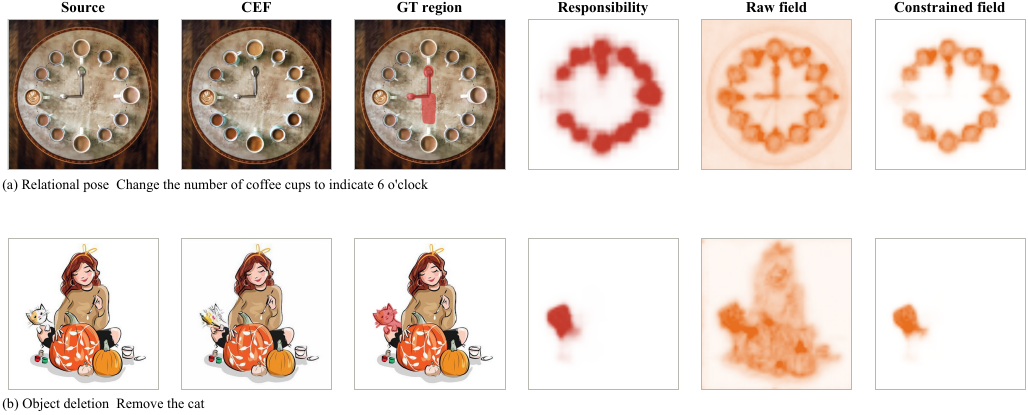}
\caption{Two representative CEF failures on SD3.5. The ground-truth (GT) edit regions are included solely as post-hoc diagnostic references. Responsibility is shown in red; raw and constrained edit-field path lengths are shown in orange with a shared scale within each example.}
\label{fig:app-failures}
\end{figure}

Figure~\ref{fig:app-failures} shows two representative failures with their post-hoc field diagnostics. In the example involving a spatial relation, the grounding phrase \emph{cups} places responsibility on the surrounding cups rather than the central clock hands identified by the benchmark edit region. Although the raw field contains a response around the clock hands, the constrained field attenuates it; the hands remain near their source pose while the cups are altered. This illustrates that noun-grounded responsibility can miss the visual carrier of a relational change.

In the deletion example, responsibility overlaps the cat and the constrained field is localized to that region. The cat is removed, but the vacated region is completed with painting tools instead of a clean continuation of the scene. This is a scene-completion ambiguity rather than a localization failure: spatial responsibility determines where the model may edit, but it does not uniquely determine content that was occluded by the removed object.

\end{document}